\documentclass[journal]{IEEEtran}
\usepackage{times}  
\usepackage{helvet} 
\usepackage{courier}  
\usepackage[hyphens]{url}  
\usepackage{graphicx} 
\usepackage{epsfig}
\usepackage{amsmath}
\usepackage{amssymb}
\usepackage{multirow}
\usepackage{makecell}
\usepackage{bm}
\usepackage{float}
\renewcommand{\figurename}{Fig.}

\title{
Pluggable Weakly-Supervised Cross-View Learning for Accurate Vehicle Re-Identification 
}

\author{Lu~Yang,
        Hongbang~Liu,
        Jinghao~Zhou,
        Lingqiao~Liu,~\IEEEmembership{Member,~IEEE,}
        Lei~Zhang,~\IEEEmembership{Member,~IEEE,}        
        Peng~Wang*,~\IEEEmembership{Member,~IEEE,}
        and~Yanning~Zhang,~\IEEEmembership{Member,~IEEE} 
}

\begin{document}

\markboth{xxxx} 
{Shell \MakeLowercase{\textit{et al.}}: Pluggable Weakly-Supervised Cross-View Learning for Accurate Vehicle Re-Identification}

\maketitle

\begin{abstract}
Learning cross-view consistent feature representation is the key for accurate vehicle Re-identification (ReID), since the visual appearance of vehicles changes significantly under different viewpoints. To this end, most existing approaches resort to the supervised cross-view learning using extensive extra viewpoints annotations, which however, is difficult to deploy in real applications due to the expensive labelling cost and the continous viewpoint variation that makes it hard to define discrete viewpoint labels. In this study, we present a pluggable Weakly-supervised Cross-View Learning (WCVL) module for vehicle ReID. Through hallucinating the cross-view samples as the hardest positive counterparts in feature domain, we can learn the consistent feature representation via minimizing the cross-view feature distance based on vehicle IDs only without using any viewpoint annotation. More importantly, the proposed method can be seamlessly plugged into most existing vehicle ReID baselines for cross-view learning without re-training the baselines. To demonstrate its efficacy, we plug the proposed method into a bunch of off-the-shelf baselines and obtain significant performance improvement on four public benchmark datasets, i.e., VeRi-776, VehicleID, VRIC and VRAI.

\end{abstract}

\begin{IEEEkeywords}
Vehicle Re-identification, Cross-view Feature, Weakly Supervised.
\end{IEEEkeywords}

\section{Introduction}

\IEEEPARstart{W}{ith} the recent widespread of video surveillance in public transportation system, vehicle re-identification (ReID) 
has become a prevalent 
computer vision application~\cite{Chen_2019_ICCV,Wang_2019_ICCV,zhao2018adversarial}, 
which aims at embedding visual appearance of various vehicles into 
an appropriate feature space where vehicles with identical IDs get gathered  
while 
ones with different IDs 
get separated with clear margins.   
However, it is still challenging to achieve accurate vehicle ReID~\cite{2019vehicle,2019VehicleRe-Identification,wang2017orientation,wei2018vp} in real applications. One of the most important reasons is the viewpoints variation problem, viz., visual appearance of a specific vehicle changes significantly under different viewpoints, which makes it difficult to learn an appropriate feature space as that mentioned above. For example, 
it has shown that vehicles with different IDs but under the same viewpoint even can obtain more similar visual appearance than that of vehicles with identical IDs but from different viewpoints
~\cite{Chu_2019_ICCV}. When casting the vehicle ReID in such a case into a deep metric learning problem with the conventional triplet loss, 
decreasing the distance between the positives (i.e., vehicles with identical IDs) from different viewpoints in the feature space will 
implicitly impede 
the negatives under the same viewpoint being 
separately with an expected margin.
Thus, it is crucial for accurate vehicle ReID to learn a cross-view feature representation. 

\begin{figure}[t]
\begin{center}
\resizebox{0.5\textwidth}{!}{
\includegraphics[width=\linewidth]{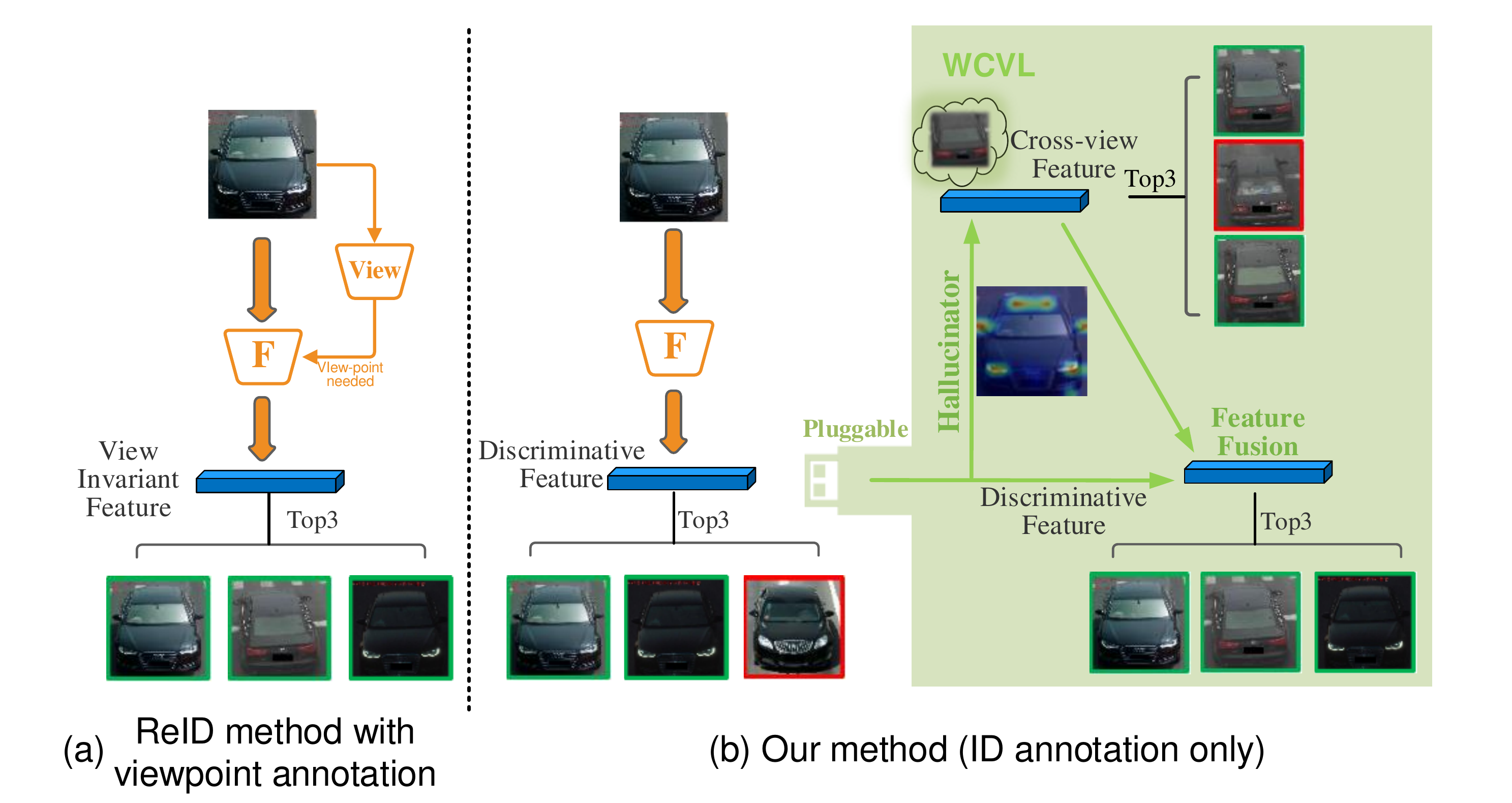}
}
\caption{Illustration of existing supervised cross-view learning method (left) and the proposed weakly supervised cross-view learning method (right).
Existing methods usually learn the view-invariant feature 
with the help of a viewpoint predictor 
that requires extensive 
viewpoint annotations for training, while the proposed method learns the cross-view feature representation through hallucinating cross-view samples as the hardest positive pairs and minimizing their distance in a specific feature space without using any extra viewpoint annotations. Moreover, it is pluggable to most existing vehicle ReID baseline.
}
\label{fig:imagination}
\end{center}
\vspace{-1em}
\end{figure}

\begin{figure*}[t]
\centering
\resizebox{0.95\textwidth}{!}{
\includegraphics[width=\linewidth]{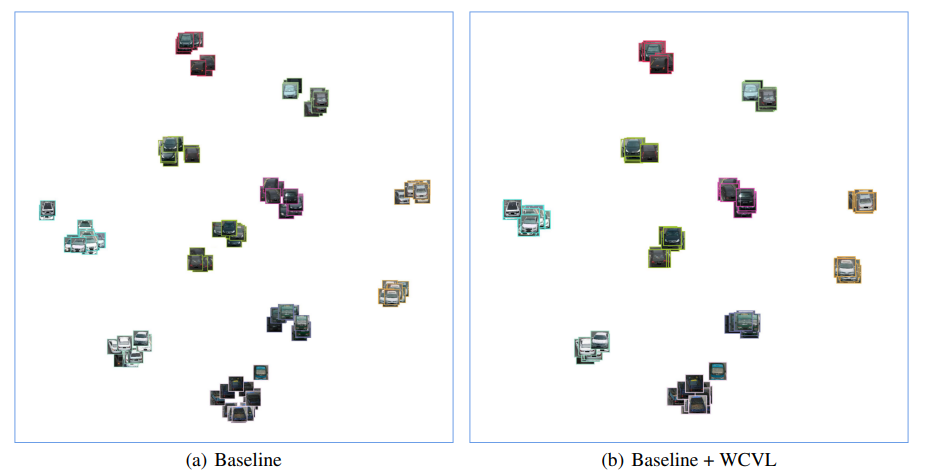}
}
\caption{Visualization of feature distribution by t-SNE on VehicleID test dataset. Different border color represents different
vehicle IDs. Images in (a) represents corresponding features from baseline and images in (b) represents corresponding features
from our approach. The baseline is trained with cross-entropy loss and hard triplet loss. We randomly chose $10$ vehicle IDs for visualization. Best viewed in color.}
\label{fig:t-SNE}
\end{figure*}

To achieve this goal, many recent works resort to the supervised cross-view learning that utilizes extra viewpoints annotations (e.g., keypoints~\cite{Wang_2017_ICCV}, viewpoints~\cite{zhou2018vehicle,Chu_2019_ICCV}) in addition to the vehicle IDs in training data to assist the feature learning with viewpoints alignment or estimation. As shown in Fig.~\ref{fig:imagination}, the extra viewpoints annotations related inference during model training empower these works to learn a view-invariant feature representation as well as obtain performance improvement.  
However, due to the expensive cost for viewpoint annotation, it is impractical to 
deploy these supervised cross-view learning into real applications.  



To mitigate this problem, we present a pluggable weakly-supervised cross-view learning method for vehicle ReID. 
Inspired by the visual behaviour of human who can hallucinate the multi-view representations for a given object from a single view, we propose to hallucinate the representation of the same vehicle under different viewpoints based on a conventional vehicle ReID dataset without using any viewpoint annotations for model training. 
Many works~\cite{Chu_2019_ICCV,bai2018group} found that large feature distance in embedding space for same ID are often caused by viewpoint variation.
As shown in Figure~\ref{fig:t-SNE}, even though the baseline method has used the hard example mining strategy, the images with the same ID will be clustered into different groups according to the viewpoint. Therefore, for a sample, the farthest positive sample is more likely to come from different viewpoint.
To this end, we consider the hardest positive pairs, i.e., two samples with the same vehicle ID but the largest distance in a pre-defined visual feature space, as samples from two different views.  Since such hallucinated cross-view samples have empirically different visual appearance but the same ID, they can be considered as a approximate alternative for the real cross-view samples with viewpoints annotations. With these hallucinated cross-view samples, we propose to minimize their distance in a latent feature domain. Apparently, when such a distance decreases to zero, we can obtain the ideal cross-view feature representation. More importantly, through implementing the cross-view learning as a separate network 
module, the proposed method can be seamlessly plugged into most exiting vehicle ReID baselines for cross-view learning without re-training the baselines. With extensive experiments on three vehicle ReID benchmark datasets, the proposed method obviously outperforms most existing non-cross-view learning and supervised cross-view learning baselines with a clear margin.

In summary, the contribution of this study is three-fold:

\begin{itemize}
\item 
We present a novel Weakly-supervised Cross-View Learning (WCVL) module for vehicle ReID, which can learn the cross-view feature representation without using any viewpoints related annotations.

\item The proposed method can be seamlessly plugged into most exiting vehicle ReID baselines for cross-view learning without re-training the baselines. 

\item We evaluated our method on three large-scale vehicle ReID benchmark datasets and obtain 
the state-of-the-art performance without using any viewpoints annotations.
\end{itemize}
\section{Related Work}
\begin{figure*}[!t]
    \begin{center}
        \includegraphics[width=0.83\textwidth]{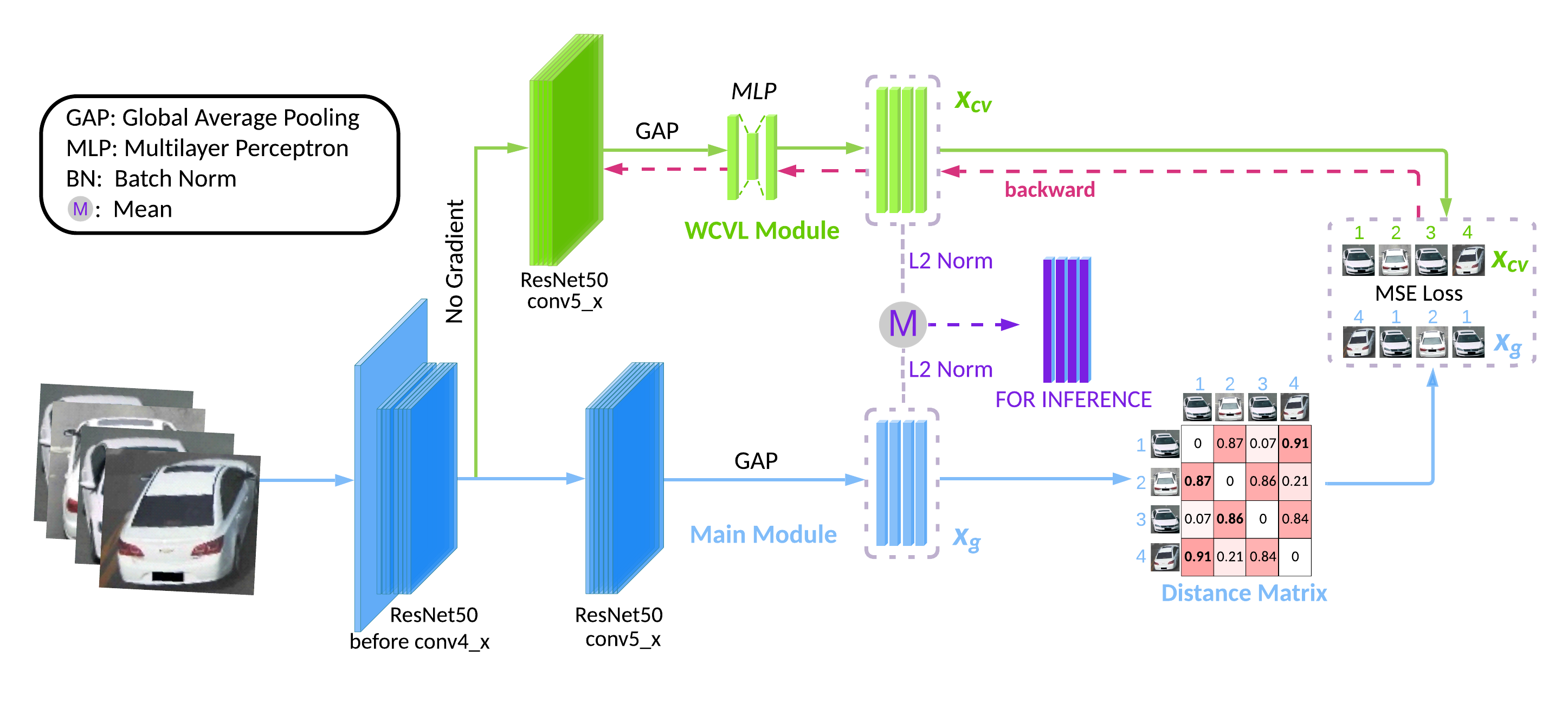}
    \end{center}

    \vspace{-1em} 
    \caption{The overall structure of our proposed model. It consists of two modules: the main module and the weakly supervised cross-view learning (WCVL) module. $x_g$ is the global (discriminative) feature and $x_{cv}$ is the cross-view feature. The main module is pre-trained with cross-entropy loss and hard triplet loss.}
    \label{fig:architecture}

\vspace{-1em} 
\end{figure*}

\subsection{Vehicle ReID}

Existing works mainly focus on casting the vehicle ReID into a deep metric learning problem~\cite{Chen_2019_ICCV} which aims at learning an embedding space for sample comparison using appropriate loss functions. To this end,
various loss functions have been customized for performance improvement.
For example, ~\cite{liu2016vehicleid} proposes a coupled clusters loss (CCL) to stabilize the training phase and accelerate the convergence speed, which extrapolates the conventional triplet in metric learning into multiple samples by measuring the distance between a given sample and a cluster center rather than a randomly selected one. Detailed, CCL measures the distances between samples and a cluster center rather than randomly selected ones, which extrapolates the triplet into multiple samples.
~\cite{zhang2017improving} proposes a classification-oriented loss to further regularize the conventional 
triplet loss.


Although these methods have gain performance improvement at some extent, few of them consider to explicitly handle viewpoints variation problem and thus show limited generalization capacity in real applications. In contrast, we present an effective cross-view feature learning approach which can well mitigate the viewpoints variation problem and obtains the state-of-the-art performance.

\subsection{Cross-View Learning}
Consider that viewpoint variation is a fundamental problem in real vehicle ReID applications, many effective solutions have been proposed recently.   
~\cite{zhou2018aware} proposes a Viewpoint-aware Attentive Multi-view Inference (VAMI) model, which extracts the single-view feature representation and then transforms it into a global multi-view one.
~\cite{Chu_2019_ICCV} introduces a viewpoint-aware metric learning scheme which first estimates the viewpoint, and then learns two metrics for similar viewpoints and different viewpoints in separate feature space.
~\cite{lou2019embedding} proposes an end-to-end embedding adversarial learning network (EALN), which generates cross-view images with GAN. The network extracts the features of two images and concatenates them as the final representation which turns out to be computation prohibitive, while our proposed WCVL module shares several features with its main module with little computation cost overhead.
It is noteworthy that all 
these approaches mentioned above 
require extra annotations such as viewpoint annotations and thus hard to generalize to most real applications without extra annotations.  
In contrast, the proposed method can conduct the cross-view learning only depending on the conventional ID annotations without using any extra viewpoint annotations. Although, a few recent works also achieve this at some extent, they are totally different from the proposed.

~\cite{bai2018group,lin2019multi} 
utilizes an online grouping method to partition samples 
with the same vehicle ID into a few groups, and then introduce 
a ranking losses to 
get samples in multiple groups closer. In contrast, the proposed method proposes to separately learn a cross-view feature and fuse it with the global (discriminative) feature learned by a vehicle ReID baseline for inference without 
clustering the samples. In addition, the proposed method can be plugged into most existing vehicle ReID baselines without re-training them while ~\cite{bai2018group,lin2019multi} fails to do this. 
\section{Approach}
\label{sec:approach}
\subsection{Weakly-supervised Cross-View Learning} 
As shown in Fig.~\ref{fig:architecture}, 
the main structure of the proposed method consists of two separate modules:
a main module (baseline) 
and 
an WCVL module.
The main module is good at retrieving images from similar viewpoints, and WCVL module is good at retrieving images from different viewpoints.
The main module employs the ResNet-50~\cite{he2016deep} pretrained on ImageNet~\cite{deng2009imagenet}, while the WCVL module further adds two full connection layers onto the end of ResNet-50~\cite{he2016deep}. 
Moreover, to decouple the feature learned by these two modules, only 
the parameters before conv4\_x in ResNet50 are shared by them. In this study, we propose to extract a global (discriminative) feature representation vector for the input vehicle image using the main module, while a generative cross-view feature representation using the WCVL module. 

To achieve this goal, we utilize a pre-trained vehicle ReID baseline without explicit cross-view learning as the main module. Most of these baselines are often trained by jointly minimizing the cross-entropy loss and the triplet loss. Specifically, for $N$ vehicle image samples selected from 
$M$ 
IDs, the cross-entropy loss can be defined as
\begin{equation}\label{Fun:CE}
L_{ce}=-\frac{1} {N}\sum^{N}_{i=1}\sum^{M}_{j=1}[y_i=j] \cdot \mathrm{log}(Prob_{i,j}),
\end{equation}
where 
$[\cdot]$ denotes the indicator function and $y_{i}$ is the ground truth 
ID for the $i$-th sample.
$Prob_{i,j}$ is the predicted probability for $i$-th sample belonging to the ID 
$j$. It has shown that cross-entropy loss is able to cluster samples with the same vehicle ID together. 
In addition, the triplet loss~\cite{schroff2015facenet} can be formulated as 
\begin{equation}
\begin{aligned}
    L_{tri}=\frac{1} {N}\sum^{N}_{i=1} \big[ \, D(\mathbf{x_i}, \mathbf{x_i}^p) - D(\mathbf{x_i}, \mathbf{x_i}^n) + \alpha  \, \big]_+,
\end{aligned}
\end{equation}
where $\mathbf{x_i}$, $\mathbf{x_i}^p$, and $\mathbf{x_i}^n$ 
denote the anchor, positive (i.e., with the same ID as the anchor) and negative (i.e., with different ID from the anchor) samples respectively. In practice, we often choose the farthest positive sample and the closest negative sample in batch to form a hard triplet. $D(\cdot)$ is a distance metric and $\alpha$ is a pre-defined margin scalar (e.g., $0.3$). 
The $[\cdot]_{+}$ denotes $max([\cdot], 0)$.
Slightly different from the cross-entropy loss, the triplet loss focuses on reducing the distance between two samples with the sample ID while enlarging the distance between two samples with different IDs. Therefore, when jointly minimizing the cross-entropy loss and the triplet loss, we can obtain a discriminative feature representation for the input vehicle image, especially when all vehicle images come from the same viewpoint. 

For the WCVL module, to yield the cross-view feature representation, we hallucinate the cross-view vehicle samples with the same ID as the hardest positive pairs, i.e., two samples with the same ID but the largest distance in a pre-defined feature space. Then, a mean squared error (MSE) loss is imposed on the WCVL module as  

\begin{equation}
\label{eq:mseloss}
L_{mse} = \frac{1} {N}\sum^{N}_{i=1} {\left \| \mathbf{x_{cv,i}} - \mathbf{x_{g,i}^p} \right \|}_{2},
\end{equation}
where $\mathbf{x_{cv}}$ 
denotes the latent cross-view feature of the anchor sample generated from the WCVL module and $\mathbf{x_g^p}$ is the 
pre-defined feature of the hardest positive sample. 
Apparently, when the MSE loss decreases to zero, we can obtain the ideal cross-view feature representation for the input vehicle image.

When both modules have been well trained, we further fuse the learned discriminative feature and the cross-view feature together for final ReID inference. 
Moreover, the decoupled WCVL module empowers us to plug the proposed method into most existing vehicle ReID baselines for cross-view learning and performance enhancement without re-training these baselines. More evidence will be provided in Section Experiments~\ref{sec:Experiments}.


\subsection{Necessity of the decoupled WCVL}
Different from most existing single-module vehicle ReID methods, the proposed method introduces an extra WCVL module and train it using a cross-view MSE loss decoupled with training the main module using the triplet loss.
To demonstrate that introducing a decoupled WCVL is better than learning a single main module using both the triplet loss and the cross-view MSE loss, we will prove that the latter scheme resembles a trivial re-weighting on the conventional triplet loss based metric learning and leads to limited performance improvement. Specifically, when the decoupled WCVL module is removed, the combined loss on the main module 
can be formulated as  

\vspace{-1em}
\begin{equation}
\label{eq:Decoupling}
\resizebox{0.49\textwidth}{!}{
$\begin{aligned}
    L_{tri + mse} &= \frac{1} {N}\sum^{N}_{i=1} (\beta - 1) D(\mathbf{x_i}, \mathbf{x_i}^p) + \big[ \, D(\mathbf{x_i}, \mathbf{x_i}^p) - D(\mathbf{x_i}, \mathbf{x_i}^n) + \alpha  \, \big]_+ \\
    &\geq \frac{1} {N}\sum^{N}_{i=1} \big[ \, \beta D(\mathbf{x_i}, \mathbf{x_i}^p) - D(\mathbf{x_i}, \mathbf{x_i}^n) + \alpha  \, \big]_+ \\
    &= \frac{1} {N}\sum^{N}_{i=1} \big[ \, \beta(D(\mathbf{x_i}, \mathbf{x_i}^p) - \triangle_{p}) - (D(\mathbf{x_i}, \mathbf{x_i}^n) - \triangle_{n})  \, \big]_+,
\end{aligned}$
}
\end{equation}
where $\triangle_{n} - \beta \triangle_{p} = \alpha$. $\triangle{n}$, $\triangle_{p}$ 
indicate the margin between two samples with different IDs and that between two samples with the same ID 
respectively. As can be seen, through exploiting the upper bound of the combined loss, the introduced cross-view MSE loss resembles re-weighting the triplet loss with a factor $\beta \geq 1$. With such a re-weighting factor,  
$\triangle_{p}$ 
is forced to be reduced $\beta$ times less and thus the 
distance between samples with the same ID will be diminished accordingly. In Section Ablation Study~\ref{sec:ablationstudy}, we will prove that such re-weighting is trivial and leads limited performance improvement.
What's more, from Figure~\ref{fig:activation} we can find that the main module and WCVL modules focus on complementary features: the main module learns robust view-specific discriminative features, while the WCVL module learns cross-view features.
Thus, the proposed decoupled WCVL is necessary for accurate vehicle ReID.


\subsection{Feature Fusion for Inference}
To fuse the discriminative feature learned by the main module and the cross-view feature learned by the WCVL module, we utilize their 
$L2$-normalized average as 
the final output feature for inference:
\begin{align}
{\mathbf{x}}= \frac{1}{2} \left ( \frac{\mathbf{{x}_{g}}}{\left \| \mathbf{{x}_{g}}\right \|}_{2} +\: \frac{\mathbf{{x}_{cv}}}{\left \| \mathbf{{x}_{cv}}\right \|}_{2} \right ),
\label{eq:norm_ave}
\end{align}

\begin{figure}[!t]
    \centering
    \resizebox{0.45\textwidth}{!}{
    \includegraphics[width=\linewidth]{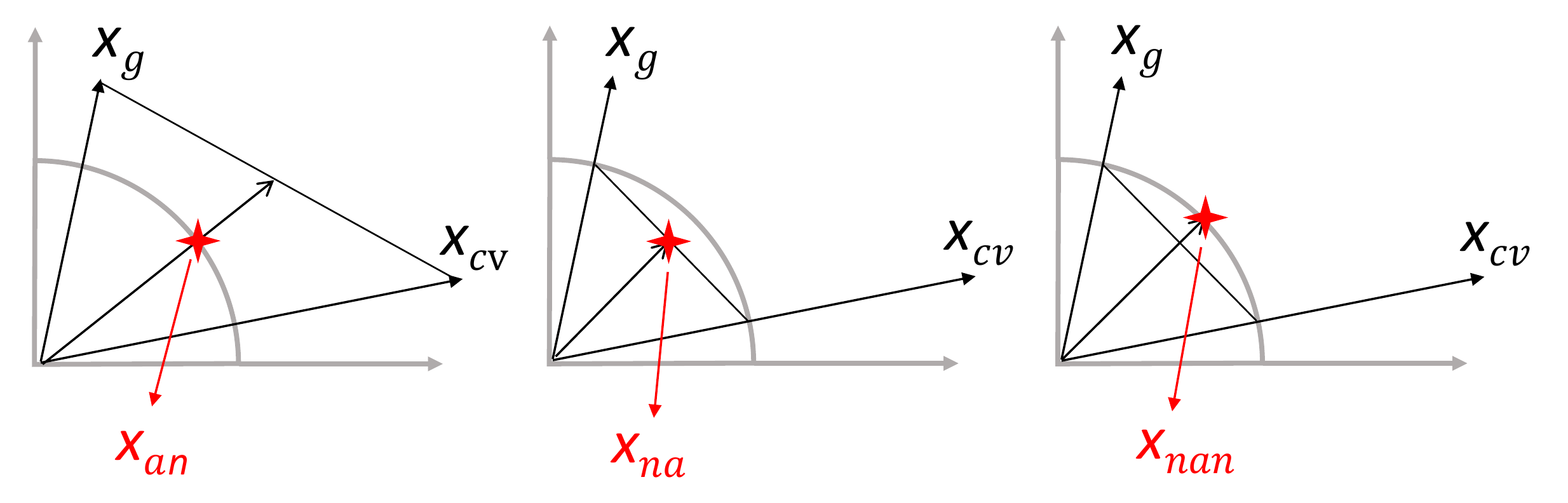}
    }
    \caption{Illustration of three different ways to normalize and average between discriminative feature (from the main module) and cross-view feature (from the WCVL module). 
    We use $\mathbf{x_{na}}$ as the final feature in the inference phase.
    }
    \label{fig:norm}
\vspace{-1em}
\end{figure}

It can be seen that such a feature fusion involves the “average” and “normalization” operations. To demonstrate the effect of the order of these two operations, we illustrate the utilized fusion scheme  $\mathbf{x_{na}}$ ($na$ is the short for normalization and average) and other two alternatives in 
Fig.~\ref{fig:norm}. 
We can find that $x_{an}$ ($an$ is short for average and normalization) is not in the middle of these two features, but closer to the feature with a larger norm.
We note that $L2$-normalization is necessary for the final representation if using cosine distance for measurement while not if using Euclidean distance. Since Euclidean distance is applied in our experiments, 
it is not necessary to further transform the average of $L2$-normalized $\mathbf{x_{na}}$  into the unit hypersphere as $\mathbf{x_{nan}}$ ($nan$ is short for normalization, average, and normalization). 
According to our experiments, $\mathbf{x_{na}}$ tends to 
obtain the best class separability.

\section{Experiments}
\label{sec:Experiments}

\subsection{Datasets}
We conduct extensive experiments on four public large-scale benchmarks for vehicle ReID. The Cumulative Match Curve (CMC) and the mean Average Precision (mAP) are used as the evaluation criteria.

\noindent{\bf{VeRi-776}}~\cite{liu2016veri-776} is a public vehicle dataset 
which consists of $49,357$ images of $776$ distinct vehicles that were captured with $20$ non-overlapping cameras in a variety of orientations and lighting conditions. 
We follow the original protocol to retrieve queries in an image-to-track fashion, where queries and the correct gallery samples must be captured from different cameras.

\noindent{\bf{VehicleID}}~\cite{liu2016vehicleid} is a widely-used vehicle ReID dataset which contains vehicle images captured in the daytime by multiple cameras. There are a total of $221,763$ images with $26,267$ identities, where each vehicle has either front or rear view. 
The training set contains $13,134$ identities while the testing set has $13,133$ identities. 
The test set is further divided into three subsets with different sizes: Small subset, medium subset, and large subset.
Noted that there is only one image for each identity in the gallery, therefore many methods~\cite{alfasly2019variational, Chu_2019_ICCV, he2019part} only use CMC as the evaluation criterion in VehicleID dataset, while others~\cite{bai2018group, lou2019veri, lou2019embedding, zhu2019vehicle, lin2019multi} still use mAP and CMC.
In order to make a comprehensive comparison with all SOTA methods, we use both mAP and CMC.

\noindent{\bf{VRIC}}~\cite{kanaci2018vric} is a more realistic and challenging vehicle ReID dataset. It is uniquely characterized by vehicle images subject to more realistic and unconstrained variations in resolution (scale), motion blur, illumination, occlusion, and viewpoint. It contains $60,430$ images of $5,622$ identities captured by $60$ different cameras from both day and night.

\noindent{\bf{VRAI}}~\cite{2019VehicleChannels} is a large-scale dataset for vehicle re-identification, which contains $137,000$ images of $13,000$ vehicle instances captured by UAV-mounted cameras. It is the largest UAV-based vehicle ReID dataset. There are various manually labelled vehicle attributes, including vehicle type, color, skylight, bumper, spare tire and luggage rack.

\begin{table*}[!t]
	\caption{Comparison with state of the art methods on VeRi-776 and VehicleID (in \%). ``*'' indicates models trained with extra annotations such as keypoints or viewpoints. ``\dag'' means by our reimplementation. The input image size is $224\times 224$ by default except the methods with ``\ddag''. \textbf{Bold} and \emph{Italic} fonts represent the best and second best performance respectively. }
	\label{tab:VeRi-776_and_VehicleID_Comparison}
	\begin{center}
		\scalebox{0.95}{
			\begin{tabular}{c|ccc|ccc|ccc|ccc}
				\hline
				\multicolumn{1}{c|}{\multirow{2}{*}{Method}} & \multicolumn{3}{c|}{VeRi-776} & \multicolumn{3}{c|}{VehicleID (Small)} & \multicolumn{3}{c|}{VehicleID (Medium)} & \multicolumn{3}{c}{VehicleID (Large)} \\ \cline{2-13} 
				\multicolumn{1}{c|}{} & \multicolumn{1}{c}{mAP} & \multicolumn{1}{c}{cmc1} & \multicolumn{1}{c|}{cmc5} & \multicolumn{1}{c}{mAP} & \multicolumn{1}{c}{cmc1} & \multicolumn{1}{c|}{cmc5} & \multicolumn{1}{c}{mAP} & \multicolumn{1}{c}{cmc1} & \multicolumn{1}{c|}{cmc5} & \multicolumn{1}{c}{mAP} & \multicolumn{1}{c}{cmc1} & \multicolumn{1}{c}{cmc5} \\
				\hline \hline
				AAVER~\cite{khorramshahi2019dual}$^*$ &$61.2$&$89.0$&$94.7$&$-$&$74.7$&$93.8$&$-$&$68.6$&$90.0$&$-$&$63.5$&$85.6$\\
				VANet~\cite{Chu_2019_ICCV}$^*$ &$66.3$&$89.8$&$96.0$&$-$&$\mathbf{88.1}$&$\mathbf{97.3}$&$-$&$\mathbf{83.2}$&$\mathbf{95.1}$&$-$&$\mathbf{80.4}$&$\mathbf{93.0}$\\
				PRND~\cite{he2019part}$^{*\ddag}$ &$74.3$&$94.3$&$98.7$&$-$&$78.4$&$92.3$&$-$&$75.0$&$88.3$&$-$&$74.2$&$86.4$\\
				\hline
				GS-TRE(ResNet50)~\cite{bai2018group} &$-$&$-$&$-$&$87.1$&$-$&$-$&$82.0$&$-$&$-$&$78.8$&$-$&$-$\\
				FDA-Net~\cite{lou2019veri} &$55.5$&$84.3$&$92.4$&$-$&$-$&$-$&$65.3$&$59.8$&$77.1$&$61.8$&$55.5$&$74.7$\\
				EALN~\cite{lou2019embedding} &$57.4$&$84.4$&$94.1$&$77.5$&$75.1$&$88.1$&$74.2$&$71.8$&$83.9$&$71.0$&$69.3$&$81.4$\\
				Mob.VFL~\cite{alfasly2019variational} &$58.1$&$87.2$&$94.6$&$-$&$73.4$&$85.5$&$-$&$69.5$&$81.0$&$-$&$67.4$&$78.5$\\
				QD-DLP~\cite{zhu2019vehicle} &$61.8$&$88.5$&$94.5$&$76.5$&$72.3$&$92.5$&$74.6$&$70.7$&$88.9$&$68.4$&$64.1$&$83.4$\\
				DMML~\cite{Chen_2019_ICCV} &$70.1$&$91.2$&$96.3$&$80.6^{\dag}$&$76.6^{\dag}$&$95.9^{\dag}$ &$77.6^{\dag}$&$73.3^{\dag}$&$93.6^{\dag}$&$72.4^{\dag}$&$67.7^{\dag}$&$89.8^{\dag}$\\
				MRL~\cite{lin2019multi}$^{\ddag}$ &$78.5$&$94.3$&$\emph{99.0}$&$\emph{87.3}$&$84.8$&$96.9$&$\emph{84.1}$&$80.9$&$94.1$&$\emph{81.2}$&$78.4$&$92.1$\\
				SAVER~\cite{ECCV2020Devil}$^{\ddag}$ &$\emph{79.6}$&$\mathbf{96.4}$&$98.6$&$-$&$79.9$&$95.2$&$-$&$77.6$&$91.1$&$-$&$75.3$&$88.3$\\
				\hline
				Baseline  &$79.3$&$94.7$&$\emph{99.0}$&$86.8$&$81.3$&$94.2$&$83.6$&$77.9$&$90.8$&$81.0$&$75.5$&$87.7$\\
				Ours (Baseline + WCVL)  &$\mathbf{80.4}$&$\emph{95.3}$&$\mathbf{99.1}$&$\mathbf{90.3}$&$\emph{85.2}$&$\emph{97.2}$&$\mathbf{87.1}$&$\emph{81.4}$&$\emph{94.6}$&$\mathbf{84.6}$&$\emph{78.6}$&$\emph{92.8}$\\
				Performance Gain  &$+1.1$&$+0.6$&$+0.1$&$+3.5$&$+3.9$&$+3.0$&$+3.5$&$+3.5$&$+3.8$&$+3.6$&$+3.1$&$+5.1$\\
				\hline
			\end{tabular}
		}
	\end{center}
\vspace{-1em}
\end{table*}

\subsection{Implementation Details}
We adopt ResNet50 as the backbone. The stride in conv5\_x is set to $1$ in the main module and set to $2$ in the WCVL module. 
All the input images were resized to $224 \times 224$ with a batch size of 64 (4 images/ID and 16 IDs). The main module is trained 120 epochs and the learning rate is initialized to $3.5\times 10^{-4}$ and divided by $10$ at the $40th$ and $70th$ epoch. The WCVL module is trained 60 epochs after the main module training is completed, and the learning rate is initialized to $3.5\times 10^{-4}$ and divided by $10$ at the $20th$ and $35th$ epoch.
We used hard triplet loss and cross-entropy loss for the main module, and MSE loss for the WCVL module. Euclidean distance is used as the distance metric and $L2$-normalization is applied to the features during the inference phase. ``Ours'' means ``Baseline + WCVL'' in the following experiments by default.

\begin{table}[!t]
\begin{center}
 \caption{Comparison with state-of-the-art methods on VRIC(in \%). \textbf{Bold} fonts represent the best performance. ``*'' refers to results represented in MSVF.}
    \label{tab:VRIC_Comparison}
\scalebox{0.98}{
\begin{tabular}{c|cc}
\hline
\multicolumn{1}{c|}{Method} & cmc1 & cmc5 \\ \hline \hline
OIFE(Single Branch)* & $24.6$ & $51.0$ \\
Siamese-Visual* & $30.5$ & $57.3$ \\
MSVF~\cite{kanaci2018vric}& $46.6$ & $65.6$ \\
CRAD~\cite{li2020cross}& $50.1$ & $68.2$ \\
BW~\cite{kumar2020strong}& $69.1$ & $90.5$ \\
\hline
Baseline                     & $74.3$ & $90.3$ \\ 
Ours (Baseline + WCVL)                  & $\mathbf{76.2}$ & $\mathbf{92.6}$ \\
Performance Gain & $+1.9$ & $+2.3$ \\
\hline
\end{tabular}
}
\end{center}
\vspace{-6mm}
\end{table}

\subsection{Comparison with State-of-the-art}

In this section we evaluate our model in comparison with state-of-the-art approaches on several benchmarks. For a fair comparison, we only demonstrate the performance of the GS-TRE model trained with ResNet-50.

For VeRi-776 and VehicleID, the results are shown in Table~\ref{tab:VeRi-776_and_VehicleID_Comparison}. 
Comparing with the approaches without extra annotation, our proposed method outperforms them on most of the evaluated settings. 
Even though MRL uses a larger image size ($256\times 256$), our method is still much better than it. 
SAVER learns instance-specific discriminative features but ignores the extreme viewpoint changes. Therefore, its performance is similar to ours on VeRi-776, but much worse than ours on VehicleID which has extreme viewpoint variation.
Comparing with the approaches with extra annotation, our model is moderately worse (less than $3\%$ on cmc1 and less than $0.5\%$ on cmc5) than VANet on VehicleID, but our method outperforms it by a large margin ($14.1\%$ on mAP and $5.5\%$ on cmc1) on VeRi-776. 
VeRi-776 dataset has continuous viewpoints while VANet uses $3$ definite viewpoints on the VeRi-776. Therefore, two instances with similar viewpoints near the viewpoint boundary may be wrongly divided into different viewpoints. Our approach uses the hardest positive sample as the different viewpoint samples, so there is no such problem.
Note that even if PRND uses extra annotation and larger image size ($512\times 512$ on VeRi-776 and $256\times 256$ on VehicleID), its performance is still much worse than our method.

For VRIC dataset, it is a newly released large vehicle dataset and there are only a few methods that have reported the results.
Our method outperforms other approaches at least $7.1\%$ on cmc1 and $2.1\%$ cmc5. Compared with baseline, we outperforms by $1.9\%$ on cmc1 and $2.3\%$ on cmc5.

For VRAI dataset, as can be seen in Table~\ref{tab:VRAI}, the proposed method outperforms all other approaches by a margin of at least $1.5\%$ on mAP. And our method is slightly better than \cite{2019VehicleChannels} which using extra annotation. As to the baseline, the method further enhance its performance by $5.7\%$, $6.9\%$ and $3.9\%$ on mAP, cmc1 and cmc5 respectively, manifesting the essential role of WCVL module.

\vspace{-3mm}

\subsection{Incorporation into other ReID Methods}
To further validate the effectiveness of our method and take advantage of its pluggable attribute, we add the WCVL module on several SOTA ReID methods~\cite{luo2019alignedreid++, Chen_2019_ICCV, dai2019batch}. Although some of these methods are originally designed for person ReID, a direct application to vehicle ReID is natural given their similar pipelines utilizing global features and the way constructing metrics. For instance, though AlignedReID~\cite{luo2019alignedreid++} is designed for person ReID, if only global features are leveraged, it can be adapted to a wide range of ReID problems.
For papers whose performance on VehicleID is unavailable, we conduct the corresponding experiments by our re-implementation. We uniform the image size from $384 \times 128$ commonly set in person ReID task to $224 \times 224$ in all the above methods for a fair comparison. As shown in Table~\ref{tab:SOTA_WCVL}, our method yields consistent performance gain over these methods in different degrees, ranging from $0.3\%$ to $2.8\%$.

\vspace{-2mm}
\begin{table}[!t]
\begin{center}
 \caption{Results on VehicleID by incorporating the WCVL module into state-of-the-art ReID methods(in \%). \textbf{Bold} fonts represent the best performance.}
 \label{tab:SOTA_WCVL}
 \scalebox{0.9}{
\begin{tabular}{c|cc|cc|cc}
\hline
{\multirow{2}{*}{VehicleID}} & \multicolumn{2}{c|}{Small} & \multicolumn{2}{c|}{Medium} & \multicolumn{2}{c}{Large} \\ \cline{2-7} 
& mAP & cmc1 & mAP & cmc1 & mAP & cmc1 \\ \hline
AlignedReID~\cite{luo2019alignedreid++} & $76.5$ & $71.7$ & $72.0$ & $67.2$ & $66.5$ & $61.5$ \\ 
AlignedReID + WCVL & $\mathbf{78.9}$ & $\mathbf{74.5}$ & $\mathbf{74.3}$ & $\mathbf{69.9}$ & $\mathbf{68.4}$ & $\mathbf{63.2}$ \\
Performance Gain & $+2.4$ & $+2.8$ & $+2.3$ & $+2.7$ & $+1.9$ & $+1.7$\\ \hline
DMML~\cite{Chen_2019_ICCV}& $80.6$ & $76.6$ & $77.6$  & $73.3$  & $72.4$  & $67.7$  \\ 
DMML + WCVL & $\mathbf{82.6}$ & $\mathbf{78.8}$ & $\mathbf{78.4}$ & $\mathbf{74.1}$ & $\mathbf{72.9}$ & $\mathbf{68.0}$ \\
Performance Gain & $+2.0$ & $+2.2$ & $+0.8$ & $+0.8$ & $+0.5$ & $+0.3$\\ \hline 
BDB~\cite{dai2019batch} & $87.1$ & $80.7$ & $84.4$ & $77.8$ & $80.8$ & $73.9$ \\ 
BDB + WCVL & $\mathbf{87.6}$ & $\mathbf{81.2}$ & $\mathbf{85.3}$ & $\mathbf{78.9}$ & $\mathbf{81.6}$ & $\mathbf{74.7}$ \\
Performance Gain & $+0.5$ & $+0.5$ & $+0.9$ & $+1.1$ & $+0.8$ & $+0.8$\\ \hline
\end{tabular}
}
\end{center}
\vspace{-1em}
\end{table}

\subsection{Ablation Study}
\label{sec:ablationstudy}
In this section, we conduct investigations on how several key factors affect the model's overall performance.

\noindent\textbf{Pluggability}
Although the scheme of end-to-end training is widely popular and often reaches optimal performance, we find in practice that cutting off the gradient from the WCVL module achieves comparable results with that of end-to-end training, shown as Table~\ref{tab:nogradient}. By discarding the back-propagation from the WCVL module, two modules are trained separately thus no interruption from each other is involved. More importantly, the observation avails us to plug our WCVL module into any off-the-shelf ReID methods without retraining thus turns it into a pluggable variant.


\begin{table}[!t]
\begin{center}
Comparative results between different models on VRAI dataset(in \%). ``*'' indicates models trained with extra annotations. \textbf{Bold} fonts represent the best performance.
\label{tab:VRAI}
\scalebox{0.98}{
\begin{tabular}{c|c|c|c}
\hline
Method &   mAP & cmc1 & cmc5 \\ \hline \hline
MGN~\cite{2018discriminative} &  $69.5$ & $67.8$ & $82.8$ \\
RAM~\cite{2018Ram} &   $69.4$ & $68.6$ & $82.3$ \\
RNN-HA~\cite{2018Coarse-to-fine} &  $74.5$ & $77.4$ & $87.4$ \\
Multi-task + DP~\cite{2019VehicleChannels}$^*$ &  $78.6$ & $\mathbf{80.3}$ & $88.5$ \\
\hline
Baseline &  $74.4$ & $72.7$ & $84.7$ \\
Ours (Baseline + WCVL) &  $\mathbf{80.1}$ & $79.6$ & $\mathbf{88.6}$\\
Performance Gain & $5.7$ & $6.9$ & $3.9$ \\
\hline
\end{tabular}
}
\end{center}
\end{table}

\begin{table}[!t]
\begin{center}
\caption{Two training schemes of the WCVL module on VehicleID. ``end-to-end'' denotes end-to-end training with the main module, while ``pluggable'' denotes the weights from the WCVL module cut off. }
\label{tab:nogradient}
\scalebox{0.98}{
\begin{tabular}{c|cc|cc|cc}
\hline
\multicolumn{1}{c|}{\multirow{2}{*}{VehicleID}} & \multicolumn{2}{c|}{Small}    & \multicolumn{2}{c|}{Medium}   & \multicolumn{2}{c}{Large}    \\ \cline{2-7} 
 & mAP & cmc1 & mAP & cmc1 & mAP & cmc1 \\ \hline \hline
Baseline & $86.8$ & $81.3$ & $83.6$ & $77.9$ & $81.0$ & $75.5$ \\ 
Ours (end-to-end)   & $\mathbf{90.3}$ & $\mathbf{85.3}$ & $\mathbf{87.1}$ & $\mathbf{81.4}$ & $\mathbf{84.6}$ & $\mathbf{78.7}$ \\ 
Ours (pluggable) & $\mathbf{90.3}$ & $85.2$ & $\mathbf{87.1}$ & $\mathbf{81.4}$ & $\mathbf{84.6}$ & $78.6$ \\ \hline
\end{tabular}
}
\end{center}
\end{table}

\noindent\textbf{Decoupled features.} 
In (\ref{eq:Decoupling}), we derive an equivalent training objective by re-weighting the within-ID and between-ID distance without decoupling features. As showcased in Table \ref{tab:Dap}, a direct combination of MSE loss and triplet loss induces little gain over the baseline and the result almost remains the same as $\beta$ varies. Therefore, the disentanglement of two conflicting requirements lies at the heart of a desirable performance while irrespective of their ratio.

\begin{table}[!t]
\begin{center}
\caption{Effect of re-weighting factor $\beta$ on VehicleID(in \%).}
\label{tab:Dap}
\scalebox{0.98}{
\begin{tabular}{c|cc|cc|cc}
\hline
\multicolumn{1}{c|}{\multirow{2}{*}{VehicleID}} & \multicolumn{2}{c|}{Small}    & \multicolumn{2}{c|}{Medium}   & \multicolumn{2}{c}{Large}    \\ \cline{2-7} 
 & mAP & cmc1 & mAP & cmc1 & mAP & cmc1 \\ \hline \hline
$\beta=1.0$   &  $86.8$ & $81.3$ & $83.6$ & $77.9$ & $81.0$ & $75.5$ \\ 
$\beta=1.5$   &  $86.8$ & $81.4$ & $83.5$ & $77.8$ & $81.0$ & $75.6$ \\ 
$\beta=2.0$   &  $\mathbf{87.0}$ & $\mathbf{81.5}$ & $\mathbf{83.8}$ & $\mathbf{78.1}$ & $\mathbf{81.1}$ & $\mathbf{75.7}$ \\ 
$\beta=6.0$   &  $86.8$ & $81.3$ & $83.4$ & $77.8$ & $80.9$ & $75.5$ \\  \hline
\end{tabular}
}
\end{center}
\end{table}

\noindent\textbf{Shared Backbone Layers.} 
In Figure~\ref{fig:architecture}, the main module and the WCVL module share layers before conv4\_x (conv1\_x~\textasciitilde~4\_x). Now we experiment with different shared layers between the main module and the WCVL module.  Experimental results in Table~\ref{tab:share_layers} show that the performance is similar for different shared layers, but the parameter size and FLOPs of the WCVL module are different.
If we share the whole ResNet50 (``conv1\_x~\textasciitilde~5\_x''), it is equivalent to use MLP in the embedding space to learn cross-view feature through discriminative feature.
Although the parameter size and FLOPs of the WCVL module drop sharply to both $2.0$Ms, its performance is lower than ``conv1\_x~\textasciitilde~4\_x'' as well. Because the discriminative feature has lost the spatial position information, it is disadvantageous for MLP to learn cross-view feature through the discriminative feature.
For better trade-off between performance and module size, we choose to share ``conv1\_x~\textasciitilde~4\_x'' in our approach.

\begin{table}[!t]
\begin{center}
 \caption{The performance with different shared layers between the main module and the WCVL module on VehicleID. 
 ``M'' is equal to $1024^{2}$, ``G'' is equal to $1024^{3}$. The parameter number of the main module is $48.1M$ and the FLOPs of the main module is $5.8G$.}
    \label{tab:share_layers}
\scalebox{0.77}{
\begin{tabular}{c|c|c|cc|cc|cc}
\hline

\multirow{2}{*}{\begin{tabular}[c]{@{}c@{}}Shared Layers\\ (conv layers)\end{tabular}} & \multirow{2}{*}{\begin{tabular}[c]{@{}c@{}}WCVL\\ \# params.\end{tabular}} & \multirow{2}{*}{\begin{tabular}[c]{@{}c@{}}WCVL\\ FLOPs\end{tabular}} & \multicolumn{2}{c|}{Small} & \multicolumn{2}{c|}{Medium} & \multicolumn{2}{c}{Large} \\ \cline{4-9} 
 &  &  & mAP & cmc1 & mAP & cmc1 & mAP & cmc1 \\ \hline \hline
1\_x~\textasciitilde~2\_x & $24.2$ M & $3.1$ G & $\mathbf{90.4}$ & $\mathbf{85.3}$ & $\mathbf{87.2}$ & $81.4$ & $84.3$ & $78.4$ \\ \hline
1\_x~\textasciitilde~3\_x & $23.0$ M & $2.1$ G & $90.2$ & $85.0$ & $\mathbf{87.2}$ & $\mathbf{81.5}$ & $\mathbf{84.8}$ & $\mathbf{78.9}$ \\ \hline
1\_x~\textasciitilde~4\_x & $16.3$ M & $0.7$ G & $90.3$ & $85.2$ & $87.1$ & $81.4$ & $84.6$ & $78.6$ \\ \hline
1\_x~\textasciitilde~5\_x & $2.0$ M & $2.0$ M & $89.5$ & $84.3$ & $86.2$ & $80.4$ & $83.8$ & $77.9$ \\ \hline
\end{tabular}
}
\end{center}
\vspace{-2mm}
\end{table}

\noindent\textbf{Normalization Approaches.} 
We conduct experiments with different normalization approaches for discriminative features and cross-view features, as shown in Figure~\ref{fig:norm}. 
We use the class separability criterion ($CSC$)~\cite{pr_book} to evaluate the effect of different normalization approaches on the class separability. It takes large values when samples in the embedding space are well clustered around their mean, within each class, and the clusters of the different classes are well separated.
The $CSC$ is calculated by between-ID scatter matrix ($S_{b}$) and within-ID scatter matrix ($S_{w}$), and is deﬁned as follows:
\begin{equation}\label{Fun:J}
\begin{aligned}
S_{b} = &\sum_{i=1}^{M}Prob_i  (\mu _i - \mu _0) (\mu _i - \mu _0)^T, \\
S_{w} = &\sum_{i=1}^{M}Prob_i   E[(x_i - \mu _i) (x_i - \mu _i)^T], \\
&CSC = \frac {trace\left\{ S_{b} \right\}} {trace\left\{ S_{w} \right\}},
\end{aligned}
\end{equation}
where $M$ is the number of classes; $Prob_i$ is the probability of class $i$; $\mu _i$ is the mean vector of class $i$, $\mu _0$ is the global mean vector; $x_i$ is the whole samples in class $i$.

In Tabel~\ref{tab:norm_ratio}, we can find that WCVL-na 
gets the highest $CSC$ among the three normalization approaches. Tabel~\ref{tab:norm_performance} shows it achieves the best performance that gives a relative mAP increases of $3.6\%$ and a cmc1 increase of $3.1\%$ on large subset of VehicleID.
\begin{table}[!t]
\begin{center}
 \caption{The $CSC$ comparison with different normalization approaches on the large subset of VehicleID.
}
\label{tab:norm_ratio}
\scalebox{0.98}{
\begin{tabular}{c|ccc}
\hline
VehicleID & $trace\left\{ S_{b} \right\}$ & $trace\left\{ S_{w} \right\}$ & $CSC$ \\ \hline \hline
Baseline & $0.680$ & $0.271$ & $2.510$ \\ 
Ours-an & $0.734$ & $0.173$ & $4.250$ \\ 
Ours-na & $0.570$ & $0.109$ & $\mathbf{5.234}$ \\ 
Ours-nan & $0.747$ & $0.144$ & $5.183$ \\ \hline
\end{tabular}
}
\end{center}
\vspace{-5mm}
\end{table}

Comparing WCVL-na and WCVL-nan in Tabel~\ref{tab:norm_performance}, we can see that $L2$-normalization for the average of $L2$-normalized features is important in cosine distance but less useful in Euclidean distance. So we do not need to further transform the average of $L2$-normalized feature on the unit hypersphere if we used Euclidean distance as the similarity measurement.

\begin{table}[!t]
\begin{center}
 \caption{The performance comparison with Euclidean distance and Dot Product distance.}
    \label{tab:norm_performance}
\scalebox{0.83}{
\begin{tabular}{c|c|cc|cc|cc}
\hline
\multicolumn{2}{c|}{\multirow{2}{*}{VehicleID}} & \multicolumn{2}{c|}{Small} & \multicolumn{2}{c|}{Medium} & \multicolumn{2}{c}{Large} \\ \cline{3-8} 
\multicolumn{2}{c|}{} & mAP & cmc1 & mAP & cmc1 & mAP & cmc1 \\ \hline \hline
\multirow{3}{*}{Euclidean} & Ours-an & $89.4$ & $84.3$ & $86.2$ & $80.4$ & $83.8$ & $78.1$ \\
 & Ours-na & $\mathbf{90.3}$ & $\mathbf{85.2}$ & $\mathbf{87.1}$ & $\mathbf{81.4}$ & $\mathbf{84.6}$ & $\mathbf{78.6}$ \\ 
 & Ours-nan & $90.1$ & $85.1$ & $86.8$ & $81.0$ & $84.4$ & $78.5$ \\ \hline
\multirow{3}{*}{Dot Product} & Ours-an & $89.4$ & $84.3$ & $86.2$ & $80.4$ & $83.8$ & $78.1$ \\
 & Ours-na & $89.5$ & $84.2$ & $86.2$ & $80.3$ & $84.0$ & $78.1$ \\
 & Ours-nan & $\mathbf{90.1}$ & $\mathbf{85.1}$ & $\mathbf{86.8}$ & $\mathbf{81.0}$ & $\mathbf{84.4}$ & $\mathbf{78.5}$ \\ \hline
\end{tabular}
}
\end{center}
\vspace{-2mm}
\end{table}

\begin{figure}[!t]
\centering
\resizebox{0.49\textwidth}{!}{
\includegraphics[width=\linewidth]{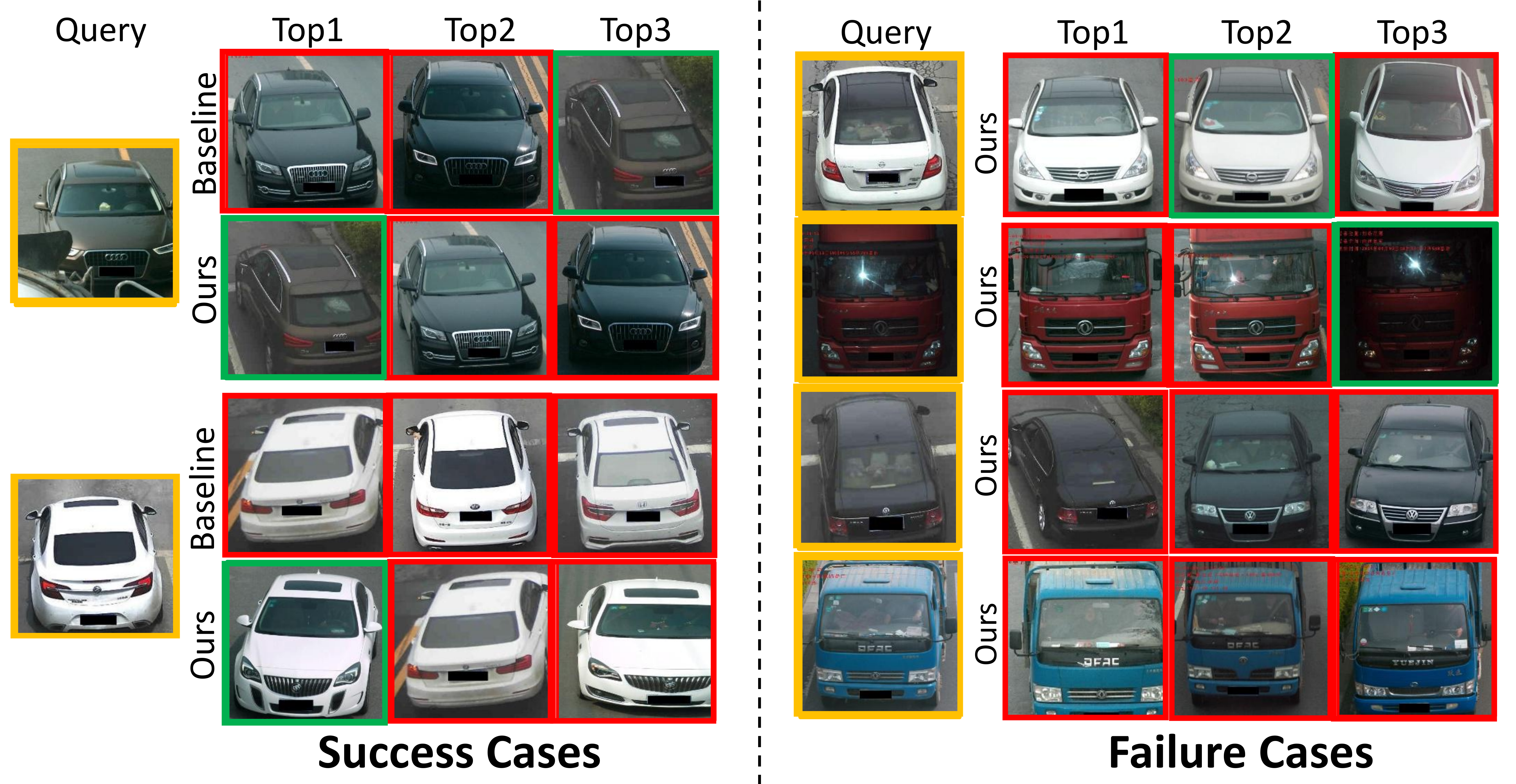}
}
\caption{Success and failure cases. In each success case, the images in the first line are from baseline and the images in the second line are from our approach. The images in failure cases are all from our approach. The images with orange border are queries. The images with a green border are positive samples, and the images with a red border are negative samples.
Best viewed in color.
}
\label{fig:FailureCases}
\vspace{-1em}
\end{figure}

\begin{figure}[!t]
	\centering
	\resizebox{0.49\textwidth}{!}{
		\includegraphics[width=\linewidth]{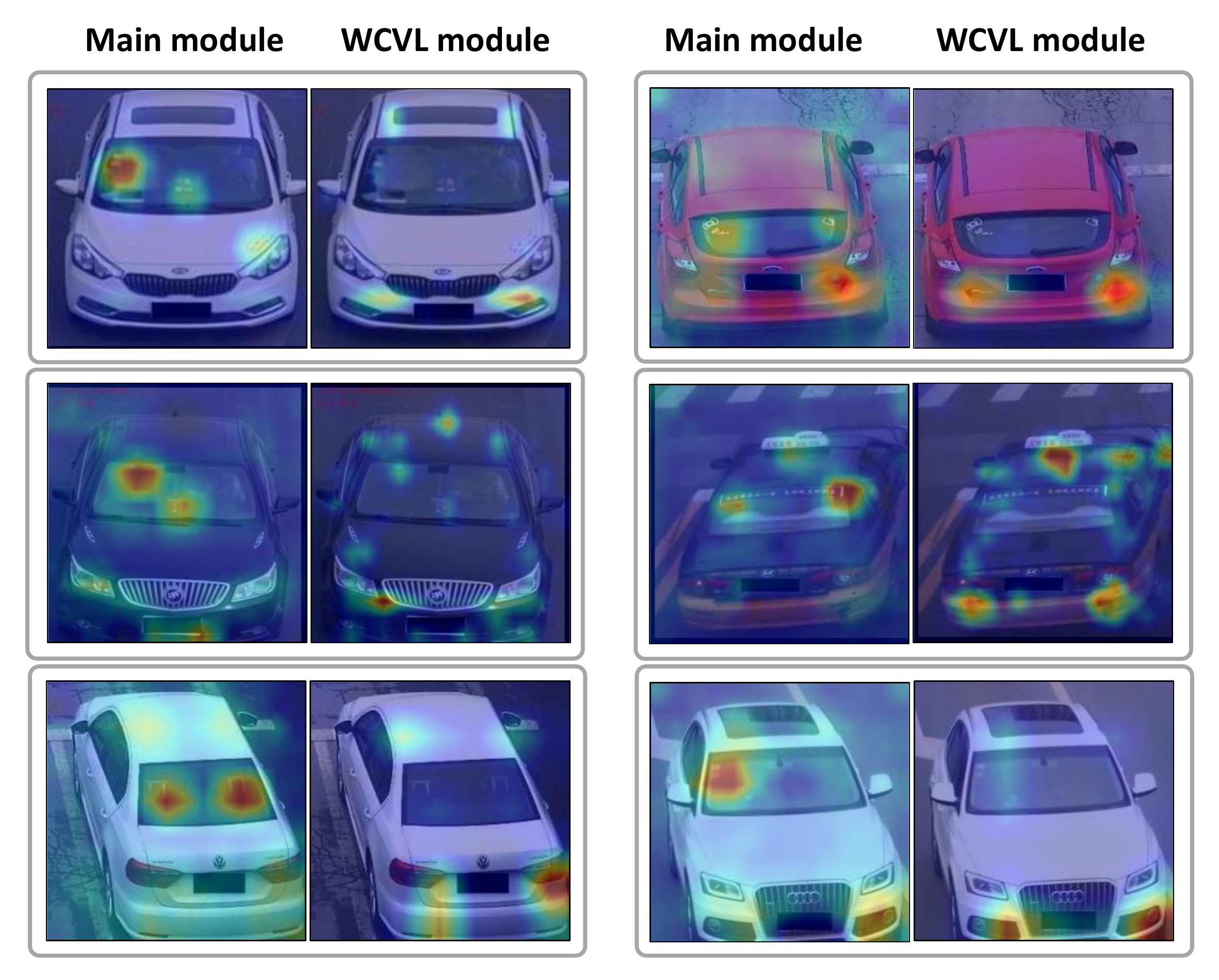}
	}
	\caption{Visualization of conv5\_x activation maps of the two modules. There are six pairs of images. In each pair of images, the left image is the activation map of the main module, and the right image is the activation map of the WCVL module.
		Best viewed in color.
	}
	\label{fig:activation}
\vspace{-2mm}
\end{figure}

\begin{figure*}[!t]
	\centering
	\includegraphics[width=0.8\linewidth]{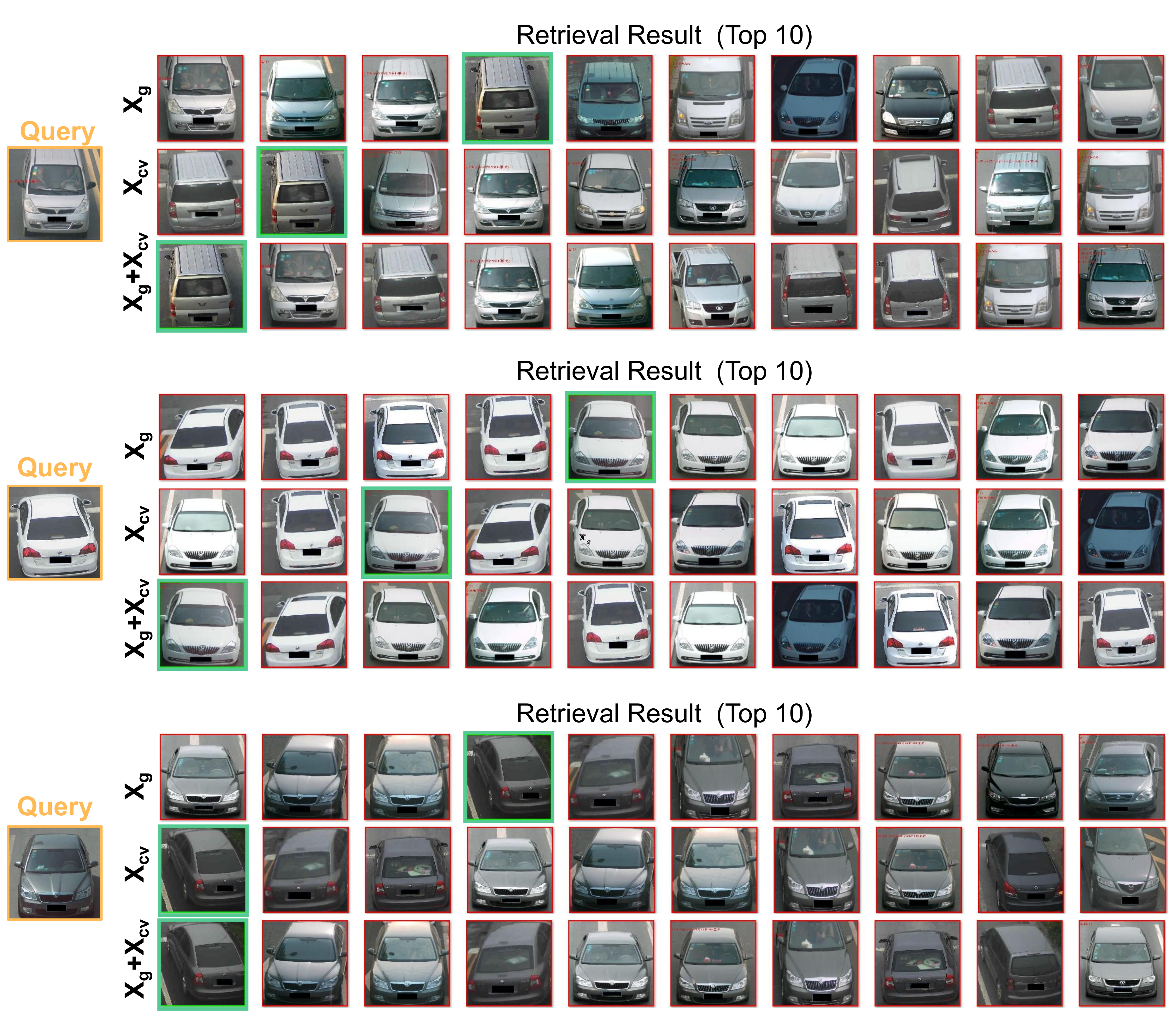}
	\caption{Top10 retrieval result with different queries. In each case, the images in the first line are retrieved by global feature $X_g$, the images in the second line are retrieved by cross-view feature $X_{cv}$ and the images in the third line are retrieved by the mean of $X_g$ and $X_cv$. The images with orange border are queries. The images with green border are positive samples, and the images with red border are negative samples. Best viewed in color.}
	\label{fig:top10}
\end{figure*}
\section{Qualitative Analysis}
In this section, we offer an insight into how the WCVL module improves ReID performance by cross-view learning.


\subsection{Success and Failure Cases}
Some success and failure cases are also presented in Figure~\ref{fig:FailureCases}. 
In the two success cases, the baseline approach cannot find out the correct positive sample with extreme viewpoint variation but chose a wrong sample from the same view-point. In such a difficult case, our approach can still get a desirable result. According to our statistics, this phenomenon accounts for $77.6\%$ of the cases where the baseline is wrong and our approach is correct. It means that the cross-view learning plays an important role in the performance gain.
Our method cannot deal with the extreme brightness variation problem well, and when the positive sample and negative sample in the gallery have the same brand, the same model, the same color and the same viewpoint.

\subsection{Activation Map Visualization}

Take the first pair of images in Figure~\ref{fig:activation} as an example, the main module focuses on view-specific discriminative parts, such as annual inspection marks, the items placed under the vehicle windshield, and the personality LOGOs on the engine hood. Conversely, the WCVL module does not pay attention to these regions 
as these features cannot generalize across different views. Instead, the WCVL focuses on regions containing view-invariant features, such as roof and lights, which are beneficial for cross-view hallucination.

\subsection{T-SNE Visualization}
Figure~\ref{fig:t-SNE} shows the feature distribution by t-SNE~\cite{maaten2008visualizing}. From Figure~\ref{fig:t-SNE}~(a) we can see that images from the
same viewpoint are easier to gather together, and the hardest
positive pairs are more likely to have different viewpoints. Based on this, we proposes to hallucinate the cross-view
samples as the hardest positive pairs and minimize their distance in a specific feature space for feature learning without
using any viewpoint annotation. From Figure~\ref{fig:t-SNE}~(b) We can see that the features obtained by our  method (Baseline + WCVL) have a more compact within-ID distribution, which is a further verification of Table~\ref{tab:norm_ratio} in the manuscript. More compact within-class distribution is useful for deep metric learning.

\subsection{Retrieval with Different Queries }
Figure~\ref{fig:top10} shows the top10 retrieval result with different query feature. We can see that the global feature are good at retrieving images from the same viewpoint, while the cross-view feature are good at retrieving images from different viewpoints. And the fusion feature (ours) of them can handle the same viewpoint and different viewpoint at the same time.

\IEEEpeerreviewmaketitle
 
\section{Conclusion}
In this work, we propose a pluggable weakly-supervised cross-view learning method to mitigate the 
viewpoints variation problem for vehicle ReID. Different from existing supervised cross-view learning methods~\cite{clark2018semi,zhu2017semi,jing2020self,wang2020deep,xiong2020cross} that requires extensive viewpoints annotations, the proposed method proposes to hallucinate the cross-view samples as the hardest positive pairs and minimize their distance in a specific feature space for feature learning without using any viewpoint annotation. Moreover, profiting from the decoupled WCVL module, the proposed method can be 
easily plugged into most exiting vehicle ReID baselines for cross-view learning without re-training the baselines.
Sufficient experiments on four benchmark vehicle ReID datasets show that
the proposed method
outperforms the state-of-the-arts by a 
clear margin, even exceeding those using extra annotations.

\ifCLASSOPTIONcaptionsoff
  \newpage
\fi

\bibliographystyle{IEEEtran}
\bibliography{IEEEabrv,main}

\end{document}